\documentclass[10pt,twocolumn,letterpaper]{article}

\usepackage{times}
\usepackage{graphicx}
\usepackage{amsmath}
\usepackage{amssymb}
\usepackage{booktabs}
\usepackage{geometry}
\usepackage{algorithm}
\usepackage{algpseudocode}
\usepackage{hyperref}
\usepackage{enumitem}
\usepackage{xspace}
\usepackage{multirow}
\usepackage{subcaption}

\newcommand{\sysname}{\textsc{AgentRunner}\xspace}

\newcommand{\eg}{\textit{e.g.}\xspace}
\newcommand{\etal}{\textit{et al.}\xspace}

\title{Beyond Autonomy: A Dynamic Tiered AgentRunner Framework\\for Governable and Resilient Enterprise AI Execution}

\author{
Kai Pan\textsuperscript{1} \and Rong Hou\textsuperscript{1}\\
\texttt{kaipan@a2alab.cn}
}

\date{}

\begin{document}

\maketitle

\begin{abstract}
The prevailing paradigm in LLM-based agent research pursues ever-greater autonomy. Yet in enterprise environments, the critical bottleneck is not insufficient autonomy but insufficient \emph{governability}: high-risk write operations proceed without independent review, complex multi-step tasks lack verification mechanisms, and indiscriminate computational expenditure renders deployment economically unviable. We present \textbf{Dynamic Tiered AgentRunner}, a controlled execution protocol distilled from a production multi-tenant SaaS platform. The framework operationalizes three core mechanisms: (1)~\emph{Risk-Adaptive Tiering} that dynamically allocates computational budget and review intensity across Light, Standard, and Full execution modes based on a task's risk-complexity profile---achieving Pareto-optimal safety-efficiency trade-offs; (2)~\emph{Separation of Powers} that physically isolates proposal (Worker), review (Critic), execution (ToolGateway), and verification (Verifier) into independent, non-colluding processes---no single agent can simultaneously propose and approve an action; and (3)~a \emph{Verifier-Recovery closed loop} that embraces failure as a first-class execution state, enabling systematic self-healing and organizational learning through retrospection. Deployed and battle-tested in a real-world multi-tenant enterprise operations platform, the framework achieves 88.9\% task success rate with 0.5\% unreviewed risk execution errors---matching always-full pipeline safety while reducing latency by 46.8\% and inference cost by 58.2\%. We argue that governability is not the antithesis of autonomy, but its prerequisite.
\end{abstract}

\section{Introduction}
\label{sec:intro}

\emph{``Enterprise AI does not suffer from a lack of autonomy, but from a lack of governability.''}

\vspace{0.5em}

The past two years have witnessed an explosion of LLM-based autonomous agent frameworks~\cite{richards2023autogpt,nakajima2023babyagi,wu2023autogen,hong2023metagpt}. These systems pursue a compelling vision: AI agents that decompose complex goals, invoke tools, and iterate toward solutions with minimal human intervention. Yet a growing body of deployment evidence reveals a fundamental tension---the very autonomy that makes these agents powerful also renders them \emph{ungovernable} in enterprise contexts.

\textbf{From Autonomy to Governability.} We observe a paradigm mismatch between the research community's pursuit of maximal autonomy and the enterprise's need for controlled, auditable, and economically sustainable execution. In production environments, three categories of failure dominate:

\begin{itemize}[leftmargin=*,nosep]
\item \textbf{Privilege Escalation.} An autonomous agent tasked with ``updating store schedules'' hallucinates a broader scope and modifies cross-brand configurations, affecting hundreds of locations without authorization.
\item \textbf{Cascading Failure.} A multi-step agent encounters a partial tool failure but continues executing downstream operations on corrupted state, propagating errors across business objects.
\item \textbf{Cost Explosion.} A static multi-agent pipeline applies full Critic-Verifier-Recovery overhead to every task---including simple read queries---multiplying inference costs by $3\text{--}5\times$ without proportional safety gains.
\end{itemize}

These are not edge cases but \emph{structural consequences} of architectures that treat governance as an afterthought. The autonomous agent paradigm (AutoGPT~\cite{richards2023autogpt}, BabyAGI~\cite{nakajima2023babyagi}) provides zero governance. Multi-agent frameworks (AutoGen~\cite{wu2023autogen}, CrewAI~\cite{crewai2024}, LangGraph~\cite{langgraph2024}) introduce role decomposition but enforce only \emph{soft constraints}---an agent ``should'' check permissions, but nothing physically prevents bypass. Traditional workflow engines (Temporal~\cite{temporal2023}, Airflow~\cite{apache2023airflow}) provide hard execution guarantees but cannot accommodate LLM reasoning.

\textbf{The AgentRunner Paradigm.} We propose a fundamental reframing: enterprise multi-agent systems should be designed not as autonomous collectives pursuing individual goals, but as \emph{governed execution ensembles}---controlled agent groups collaborating around a single business objective under explicit constitutional constraints. The key insight is that \emph{not all tasks deserve equal governance overhead}. A simple information query and a cross-brand batch mutation have fundamentally different risk profiles and should receive proportionally different levels of scrutiny.

This insight leads to our framework: \textbf{Dynamic Tiered AgentRunner}---a risk-adaptive, multi-role execution architecture that dynamically adjusts governance intensity to match task risk, enforces physical separation between proposal and execution, and builds resilience through systematic failure handling.

\textbf{Contributions.} This paper makes four contributions:

\begin{itemize}[leftmargin=*,nosep]
\item A \emph{Risk-Adaptive Tiering} mechanism that achieves Pareto-optimal trade-off between safety and efficiency, routing 55\% of production tasks through a minimal-overhead Light path while reserving full governance for the 15\% that genuinely require it.
\item A \emph{Separation of Powers} architecture with physically isolated execution boundaries---the ToolGateway acts as a hard constitutional constraint, not a prompt-level suggestion.
\item \emph{Resilience-by-Design} through a Verifier-Recovery closed loop that treats failure as a first-class state, achieving 67\% automated recovery rate on initially-failed tasks.
\item \emph{Production deployment evidence} from a real-world multi-tenant SaaS platform, demonstrating that the framework transforms AI governance from a ``prompt engineering hope'' into a ``system architecture guarantee.''
\end{itemize}

\section{The Governability Gap in Multi-Agent Systems}
\label{sec:gap}

Existing multi-agent frameworks excel at orchestration but fundamentally lack \emph{governance}. We define the governability gap as the absence of system-level mechanisms that prevent, detect, and recover from agent misbehavior independent of the underlying LLM's alignment.

\subsection{Soft Constraints Are Not Governance}

AutoGen~\cite{wu2023autogen} provides flexible multi-agent conversation patterns with customizable termination conditions. CrewAI~\cite{crewai2024} formalizes role-based task decomposition with configurable delegation. LangGraph~\cite{langgraph2024} enables graph-based agent orchestration with conditional routing. MetaGPT~\cite{hong2023metagpt} and ChatDev~\cite{qian2023chatdev} demonstrate impressive role-playing for software engineering.

These are excellent \emph{orchestration} tools. But they operate under a critical assumption: \textbf{agents are well-behaved by construction}. Their ``constraints'' are prompt-level instructions (``You are a careful reviewer...'') or conversation-level patterns (``Agent B reviews Agent A's output''). Nothing in their architecture \emph{physically prevents} a Worker agent from bypassing review and directly invoking a destructive tool. The constraint lives in the prompt, not in the execution boundary.

\subsection{Three Enterprise Disasters}

We formalize three disaster categories that soft-constraint frameworks cannot prevent:

\textbf{Disaster 1: Privilege Escalation.} An agent instructed to ``update training materials for Brand A'' generates tool calls that modify Brand B's resources. In prompt-constrained systems, scope enforcement depends on the LLM correctly interpreting boundary instructions---a fundamentally unreliable mechanism given hallucination rates.

\textbf{Disaster 2: Cascading Failure.} An agent executing a multi-step plan encounters a tool failure at step 3 of 7. Without a verification mechanism, it proceeds with steps 4--7 operating on incomplete or corrupted state. The downstream damage far exceeds the original failure.

\textbf{Disaster 3: Cost Explosion.} A static multi-agent pipeline applies identical overhead to every task. When 55\% of production tasks are simple read queries, the $3\text{--}5\times$ cost multiplier of full pipeline execution renders the system economically unviable at scale.

\subsection{Position of This Work}

Traditional workflow systems (Temporal~\cite{temporal2023}, Prefect~\cite{prefect2024}, Airflow~\cite{apache2023airflow}) solve durability and retry but cannot accommodate non-deterministic LLM reasoning. Agent safety research (Constitutional AI~\cite{bai2022constitutional}, ToolEmu~\cite{ruan2024toolemu}, R-Judge~\cite{yuan2024rjudge}) evaluates safety at the individual call level but does not provide system-level execution governance. Reflexion~\cite{shinn2023reflexion} and LATS~\cite{zhou2024lats} introduce self-reflection but within single-agent loops without separation of powers.

\sysname provides a \emph{model-agnostic governance layer}: regardless of which LLM serves as Worker or Critic, the ToolGateway's hard constraints, the Checkpoint's durability guarantees, and the tier routing logic remain invariant. The governance does not degrade when models change.

\section{Core Principles}
\label{sec:principles}

We formalize three design principles that distinguish \sysname from prior multi-agent architectures.

\subsection{Principle 1: Risk-Adaptive Tiering}

Not every task deserves three rounds of GPT-4 review. We operationalize this observation through a tier function that selects the minimal governance intensity sufficient for a task's risk profile:

\begin{equation}
\tau(T) = \arg\min_{t \in \{L, S, F\}} \textit{Cost}(t) \quad \text{s.t.} \quad \textit{Safety}(t) \geq \textit{Risk}(T)
\label{eq:tier-principle}
\end{equation}

The three tiers represent Pareto-optimal points on the safety-efficiency frontier:
\begin{itemize}[leftmargin=*,nosep]
\item \textbf{Light ($L$):} Orchestrator + Worker + ToolGateway. For read-only queries and low-risk single operations. Sub-10-second latency, minimal cost.
\item \textbf{Standard ($S$):} Adds independent CriticAgent. For write operations and multi-step tasks requiring pre-execution review.
\item \textbf{Full ($F$):} Adds Verifier + Recovery. For batch mutations, cross-domain operations, and high-impact tasks requiring post-execution validation and automated repair.
\end{itemize}

The critical property is that tier selection is \emph{dynamic}: a task initially classified as Light may escalate to Standard mid-execution if write operations are detected, and to Full if cross-domain scope is identified. This escalation is unidirectional and conservative---once elevated, a tier cannot be demoted by the Worker agent alone.

\subsection{Principle 2: Separation of Powers}

We draw an explicit analogy to constitutional governance. In our architecture:
\begin{itemize}[leftmargin=*,nosep]
\item The \textbf{Worker} proposes actions (legislative).
\item The \textbf{Critic} reviews and may veto proposals (judicial).
\item The \textbf{ToolGateway} executes approved actions under strict constraints (executive).
\item The \textbf{Verifier} validates outcomes against criteria (audit).
\end{itemize}

These roles execute as \emph{physically separate processes} with independent LLM calls and distinct prompt configurations. No single process can simultaneously propose an action, approve it, and execute it. The ToolGateway is not a suggestion---it is the \emph{only} path through which any agent role can affect the external world. Bypassing it is architecturally impossible, not merely discouraged.

\subsection{Principle 3: Resilience by Design}

We abandon the ``first-try success'' assumption. In production enterprise environments, partial failures, ambiguous results, and unexpected scope are \emph{normal operating conditions}. The framework explicitly models failure as a first-class execution state:

\begin{itemize}[leftmargin=*,nosep]
\item The \textbf{Verifier} may output \texttt{incomplete}, \texttt{failed}, or \texttt{uncertain}---not just \texttt{passed}.
\item The \textbf{Recovery} agent generates repair paths while maintaining an avoidance list of previously-failed approaches.
\item The \textbf{Retrospector} extracts organizational learning from both successes and failures, generating reusable skill drafts.
\end{itemize}

This transforms the system from a fragile single-attempt executor into a resilient closed-loop controller that improves over time.

\section{The AgentRunner Architecture}
\label{sec:arch}

\subsection{Orchestrator and Tier Routing}
\label{sec:orchestrator}

The OrchestratorAgent serves as the entry controller for every task. It performs intent classification, risk assessment, tier selection, and role activation. Formally, given task $T = (g, \mathcal{C}, \mathcal{K})$ with goal $g$, constraints $\mathcal{C}$, and context $\mathcal{K}$, the Orchestrator produces:

\begin{equation}
O(T) = (\tau, \mathcal{R}_{active}, \mathbf{sc}, \phi_0)
\end{equation}

where $\tau \in \{L, S, F\}$ is the selected tier, $\mathcal{R}_{active} \subseteq \mathcal{R}$ is the activated role set, $\mathbf{sc}$ is the success criteria vector, and $\phi_0$ is the initial phase.

\textbf{Tier Selection Logic.} The tier is determined by:
\begin{equation}
\tau(T) = \begin{cases}
L & \text{if } R(T) \leq \theta_L \;\wedge\; \neg\textit{write}(T) \;\wedge\; |\textit{scope}(T)| \leq 1 \\
S & \text{if } R(T) \leq \theta_S \;\vee\; \textit{write}(T) \\
F & \text{otherwise}
\end{cases}
\label{eq:tier-select}
\end{equation}

where $R(T)$ is a composite risk score. We define $R(T)$ as a weighted heuristic integrating four operationalized risk signals:
\begin{equation}
R(T) = w_1 \cdot \textit{op\_type}(T) + w_2 \cdot \textit{obj\_count}(T) + w_3 \cdot \textit{cross\_domain}(T) + w_4 \cdot \textit{hist\_fail}(T)
\label{eq:risk}
\end{equation}
where $\textit{op\_type} \in \{0\text{:read},\; 0.3\text{:single-write},\; 0.7\text{:batch-write},\; 1.0\text{:delete/irreversible}\}$ encodes operation severity; $\textit{obj\_count} \in [0,1]$ is the normalized count of affected business objects; $\textit{cross\_domain} \in \{0,1\}$ indicates whether the task spans multiple organizational boundaries (brands, locations); and $\textit{hist\_fail} \in [0,1]$ is the empirical failure rate of similar tasks over the preceding 30-day window. Weights $(w_1{=}0.35, w_2{=}0.25, w_3{=}0.25, w_4{=}0.15)$ are calibrated via a small-scale annotation study ($n$=120 tasks labeled by domain experts) and held fixed across tenants. The thresholds $\theta_L{=}0.25$ and $\theta_S{=}0.60$ are similarly calibrated. We note that this heuristic formulation is deliberately simple and interpretable; Section~\ref{sec:discussion} discusses potential evolution toward a lightweight learned classifier.

\textbf{Escalation Conditions.} Mid-execution escalation triggers include: (a)~Worker selecting write tools in a Light-tier task; (b)~Critic identifying undisclosed risk factors; (c)~scope expansion to multiple entities or cross-boundary operations; (d)~ToolGateway risk assessment returning medium/high. Escalation is strictly monotonic: $L \rightarrow S \rightarrow F$.

\textbf{Demotion Constraints.} Only the Orchestrator may demote a tier, and only when the elevated risk context has been explicitly resolved. The Worker cannot self-certify reduced risk. This asymmetry reflects the conservative principle: escalation is cheap (safety improves), demotion is dangerous (safety may degrade).

\subsection{Multi-Role Pipeline and State Machine}
\label{sec:pipeline}

The execution proceeds through a well-defined phase state machine:
\begin{equation}
\textit{Planning} \rightarrow \textit{Criticizing} \rightarrow \textit{Executing} \rightarrow \textit{Verifying} \rightarrow \{\textit{Recovering} | \textit{Finalizing}\} \rightarrow \textit{Retrospecting}
\label{eq:phases}
\end{equation}

Not all phases are active for all tiers: Light skips Criticizing and Verifying; Standard skips Verifying (unless critical write).

\textbf{OrchestratorAgent.} \emph{Purpose:} Intent understanding, tier determination, phase control, and inter-role arbitration. \emph{Constraints:} Cannot execute tools. Cannot override system-level hard rules. Must justify tier selections.

\textbf{WorkerAgent.} \emph{Purpose:} Plan generation and tool call intent formulation. \emph{Protocol:} Outputs structured plans with explicit assumptions, risk annotations, and user-input flags. In Light tier, intents proceed directly to ToolGateway. In Standard/Full, intents route to Critic first. \emph{Constraints:} Cannot self-approve write operations in Standard/Full tiers.

\textbf{CriticAgent.} \emph{Purpose:} Independent pre-execution review through a separate, more conservative LLM call. \emph{Protocol:} Evaluates scope boundaries, permission alignment, missing steps, and risk factors. Produces verdict $v \in \{\textit{approve}, \textit{revise}, \textit{reject}, \textit{ask\_user}, \textit{escalate}\}$. \emph{Constraints:} Cannot execute tools. Cannot override Orchestrator's tier decision (can only request escalation). Uses distinct prompt temperature and system instructions from Worker.

\textbf{VerifierAgent.} \emph{Purpose:} Post-execution validation against predetermined success criteria. \emph{Protocol:} Assesses result completeness, object integrity, and business rule compliance. Outputs status $s \in \{\textit{passed}, \textit{incomplete}, \textit{failed}, \textit{uncertain}\}$ with itemized evidence. \emph{Constraints:} Cannot execute tools. Cannot mark \texttt{uncertain} as \texttt{passed}.

\textbf{RecoveryAgent.} \emph{Purpose:} Failure analysis and repair path generation. \emph{Protocol:} Maintains avoidance list of failed payloads. Proposes $d \in \{\textit{retry}, \textit{replan}, \textit{ask\_user}, \textit{wait}, \textit{fail}\}$. Repair proposals re-enter the pipeline through Orchestrator. \emph{Constraints:} Cannot execute tools directly. Modified high-risk payloads must re-enter Critic review. Bounded by retry budget.

\textbf{RetrospectorAgent.} \emph{Purpose:} Asynchronous post-task analysis for organizational learning. \emph{Protocol:} Generates success/failure pattern summaries, outcome memories, and skill draft candidates. \emph{Constraints:} Executes asynchronously, never blocks the user-facing path. Skill drafts default to draft status requiring human approval.

\subsection{ToolGateway: The Hard Constitution}
\label{sec:gateway}

The ToolGateway is the architectural keystone of our governance model. It serves as the \emph{sole physical interface} between agent intent and real-world side effects. Any attempt to bypass it---whether by prompt injection, hallucinated direct access, or role confusion---is architecturally impossible, not merely prohibited by convention.

\textbf{Six-Layer Validation Pipeline.} Every tool invocation traverses:
\begin{enumerate}[leftmargin=*,nosep]
\item \textbf{Schema:} Structural and type validation of all parameters.
\item \textbf{Permission:} RBAC enforcement against the initiating user's permission set.
\item \textbf{Scope:} Tenant, brand, and location boundary verification---prevents cross-tenant access regardless of LLM output.
\item \textbf{Risk:} Dynamic risk scoring; medium/high triggers human confirmation workflow that \emph{physically halts} execution.
\item \textbf{Idempotency:} Duplicate detection via task-bound idempotency keys; prevents repeated mutations during recovery/retry.
\item \textbf{Execution:} Actual tool invocation with structured result capture and audit logging.
\end{enumerate}

\textbf{Agent-First Tool Protocol.} Unlike traditional REST APIs designed for human-operated UIs, tools exposed through the Gateway follow an agent-optimized protocol:
\begin{itemize}[leftmargin=*,nosep]
\item \emph{Semantic inputs:} Accept natural language descriptions alongside exact identifiers, supporting the ambiguity inherent in LLM-generated requests.
\item \emph{Structured outputs:} Return confidence scores, ambiguity indicators, evidence references, and actionable next-step suggestions.
\item \emph{Recoverable errors:} Provide machine-readable error codes (\texttt{ambiguous\_query}, \texttt{scope\_violation}, \texttt{idempotency\_conflict}), candidate resolutions, and explicit retry eligibility.
\item \emph{Dry-run support:} Preview execution without side effects, enabling Critic assessment of actual outcomes before commitment.
\end{itemize}

\textbf{Constitutional Metaphor.} We deliberately frame the ToolGateway as a constitution rather than a guideline. In multi-agent systems with soft constraints, governance degrades under adversarial conditions (prompt injection, model degradation, hallucination spikes). The ToolGateway's guarantees are \emph{model-agnostic}: they hold regardless of which LLM generates the tool call, because enforcement occurs at the system layer, not the prompt layer.

\subsection{Checkpoint and Execution Durability}
\label{sec:checkpoint}

The Runner maintains a persistent checkpoint structure:
\begin{equation}
\textit{CP} = (\tau, \phi, \mathcal{R}_{active}, \mathbf{O}, \mathbf{V}, \mathbf{Rc}, \mathbf{Rt})
\end{equation}
where $\mathbf{O}$ is the ordered list of agent opinions, $\mathbf{V}$ the verification state, $\mathbf{Rc}$ recovery history, and $\mathbf{Rt}$ retrospective output.

The checkpoint serves as the authoritative state source. Upon task suspension (human approval pending), system restart, or explicit resumption, the Runner reconstructs exclusively from checkpoint---no ephemeral in-memory state survives restarts.

\textbf{Event Protocol.} Structured events are emitted at each state transition: \texttt{runner.tier.selected}, \texttt{agent.critic.reviewed}, \texttt{agent.verifier.checked}, \texttt{runner.phase.changed}, \texttt{runner.completed}, \texttt{runner.failed}. These enable real-time observability dashboards, offline audit replay, and integration with external monitoring systems.

\subsection{Execution Flow Algorithm}
\label{sec:algorithm}

Algorithm~\ref{alg:runner} presents the unified execution logic spanning all three tiers.

\begin{algorithm}[t]
\caption{Dynamic Tiered AgentRunner Execution}
\label{alg:runner}
\small
\begin{algorithmic}[1]
\Require Task $T = (g, \mathcal{C}, \mathcal{K})$, budget $B$
\Ensure Result $r$, Trace $\Gamma$
\State $\tau, \mathcal{R}_{active}, \mathbf{sc} \leftarrow \text{Orchestrator}(T)$
\State $\textit{plan} \leftarrow \text{Worker}(T, \mathcal{K}, \mathbf{sc})$
\State $\textit{rounds} \leftarrow 0$
\If{$\tau \in \{S, F\}$} \Comment{Separation of Powers}
    \Repeat
        \State $v \leftarrow \text{Critic}(\textit{plan}, \mathbf{sc})$
        \If{$v.\textit{verdict} = \textit{escalate}$}
            \State $\tau \leftarrow \text{Orchestrator.escalate}(\tau)$
        \ElsIf{$v.\textit{verdict} = \textit{revise}$}
            \State $\textit{plan} \leftarrow \text{Worker.revise}(\textit{plan}, v)$
        \EndIf
        \State $\textit{rounds} \leftarrow \textit{rounds} + 1$
    \Until{$v.\textit{verdict} = \textit{approve}$ \textbf{or} $\textit{rounds} \geq B_{\textit{critic}}$}
    \If{$v.\textit{verdict} \notin \{\textit{approve}\}$}
        \State \Return $\text{blocked}(v)$
    \EndIf
\EndIf
\State $r \leftarrow \text{ToolGateway.execute}(\textit{plan}, \tau)$ \Comment{Hard boundary}
\If{$\tau = F$} \Comment{Resilience loop}
    \State $s \leftarrow \text{Verifier}(r, \mathbf{sc})$
    \While{$s.\textit{status} \neq \textit{passed}$ \textbf{and} $\textit{rounds} < B_{\textit{recovery}}$}
        \State $d \leftarrow \text{Recovery}(s, r, \textit{plan})$
        \If{$d.\textit{decision} = \textit{fail}$} \State \textbf{break} \EndIf
        \State $\textit{plan} \leftarrow d.\textit{repair\_plan}$
        \State $r \leftarrow \text{ToolGateway.execute}(\textit{plan}, \tau)$
        \State $s \leftarrow \text{Verifier}(r, \mathbf{sc})$
        \State $\textit{rounds} \leftarrow \textit{rounds} + 1$
    \EndWhile
\EndIf
\State $\text{Checkpoint.persist}(\tau, \phi, r, \Gamma)$
\State $\text{async:}~\text{Retrospector}(\Gamma)$
\State \Return $r$
\end{algorithmic}
\end{algorithm}

\textbf{Light Path} (lines 1--2, 16, 22--23): Bypasses Critic and Verifier entirely. Single Orchestrator+Worker call followed by ToolGateway execution. Achieves median 8.4s latency.

\textbf{Standard Path} (lines 4--15, 16, 22--23): Includes bounded Critic review loop. Worker revises until Critic approves or budget exhausts. Critic may trigger tier escalation.

\textbf{Full Path} (lines 4--15, 16--21, 22--23): Adds Verifier-Recovery loop post-execution. Bounded by recovery budget with circuit-breaker semantics.

\section{Production Evidence}
\label{sec:production}

The \sysname framework is deployed and operational in a production enterprise platform. This section presents direct evidence from the deployed system.

\subsection{Deployment Context}

The framework operates within a multi-tenant SaaS platform for enterprise chain operations management. The platform serves multiple enterprise tenants across distinct brands and locations, handling operational tasks including resource management, training program generation, schedule coordination, and operational analytics. The system supports concurrent Runner instances with model-agnostic LLM backends (MiniMax-M2.7, Kimi-K2.6) configurable per-role and per-tenant.

\subsection{Execution Trace Evidence}

\begin{figure}[t]
\centering
\includegraphics[width=\columnwidth]{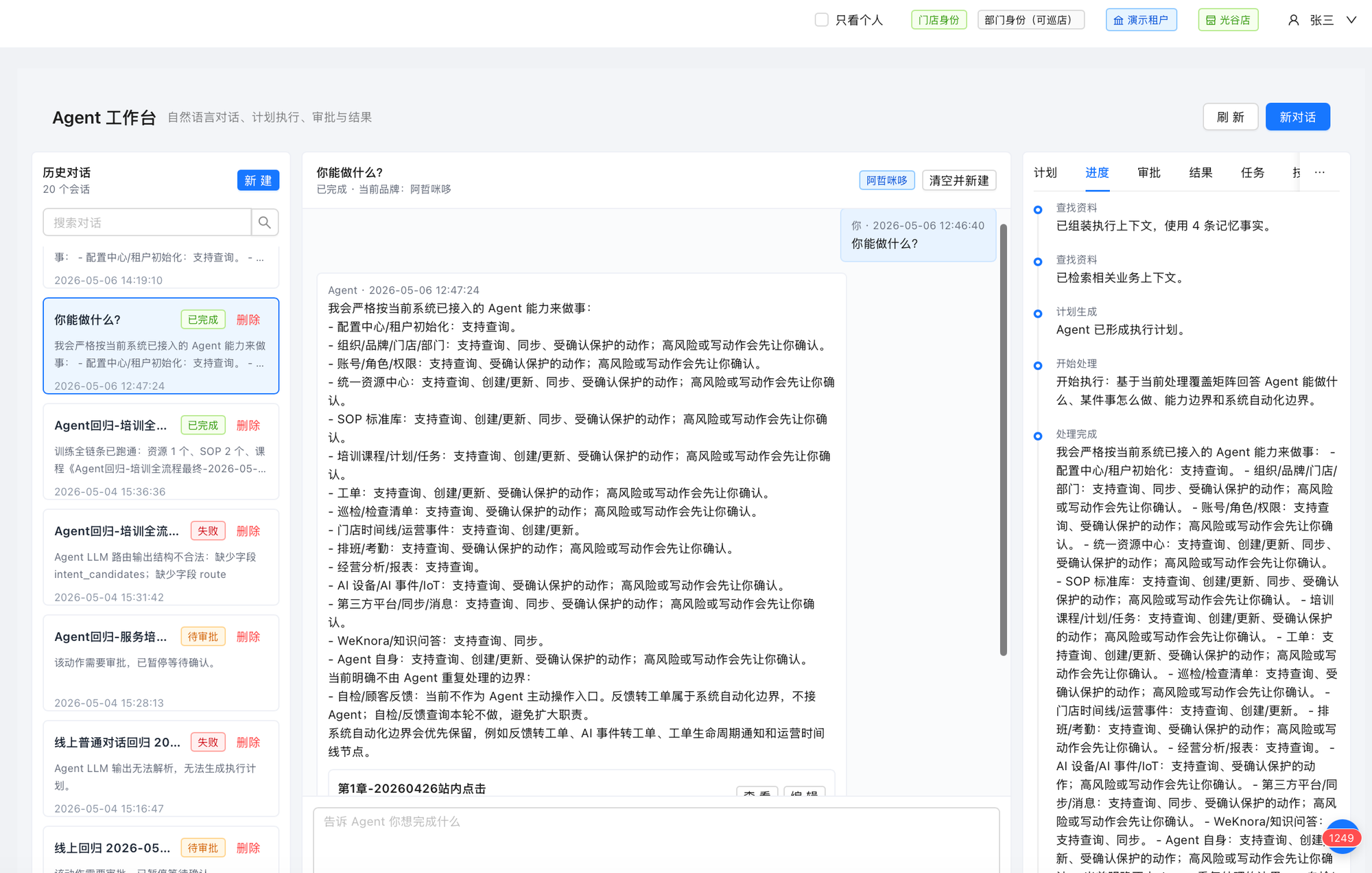}
\caption{Phase Trace of a Standard Runner in production. The right panel shows Orchestrator-driven phase transitions (intent recognition $\rightarrow$ plan generation $\rightarrow$ execution $\rightarrow$ completion). The left panel preserves explicit failure records as first-class entries, confirming the resilience-by-design principle---failures are visible audit artifacts, not hidden exceptions.}
\label{fig:phase-trace}
\end{figure}

Figure~\ref{fig:phase-trace} demonstrates the system's execution trace in production. Each Runner phase transition is captured as a persistent, auditable event. Critically, the task history panel displays failed tasks with equal prominence to successful ones---failures are not suppressed or hidden, but treated as first-class citizens in the system's state machine.

\subsection{Human-in-the-Loop Governance Evidence}

\begin{figure}[t]
\centering
\includegraphics[width=\columnwidth]{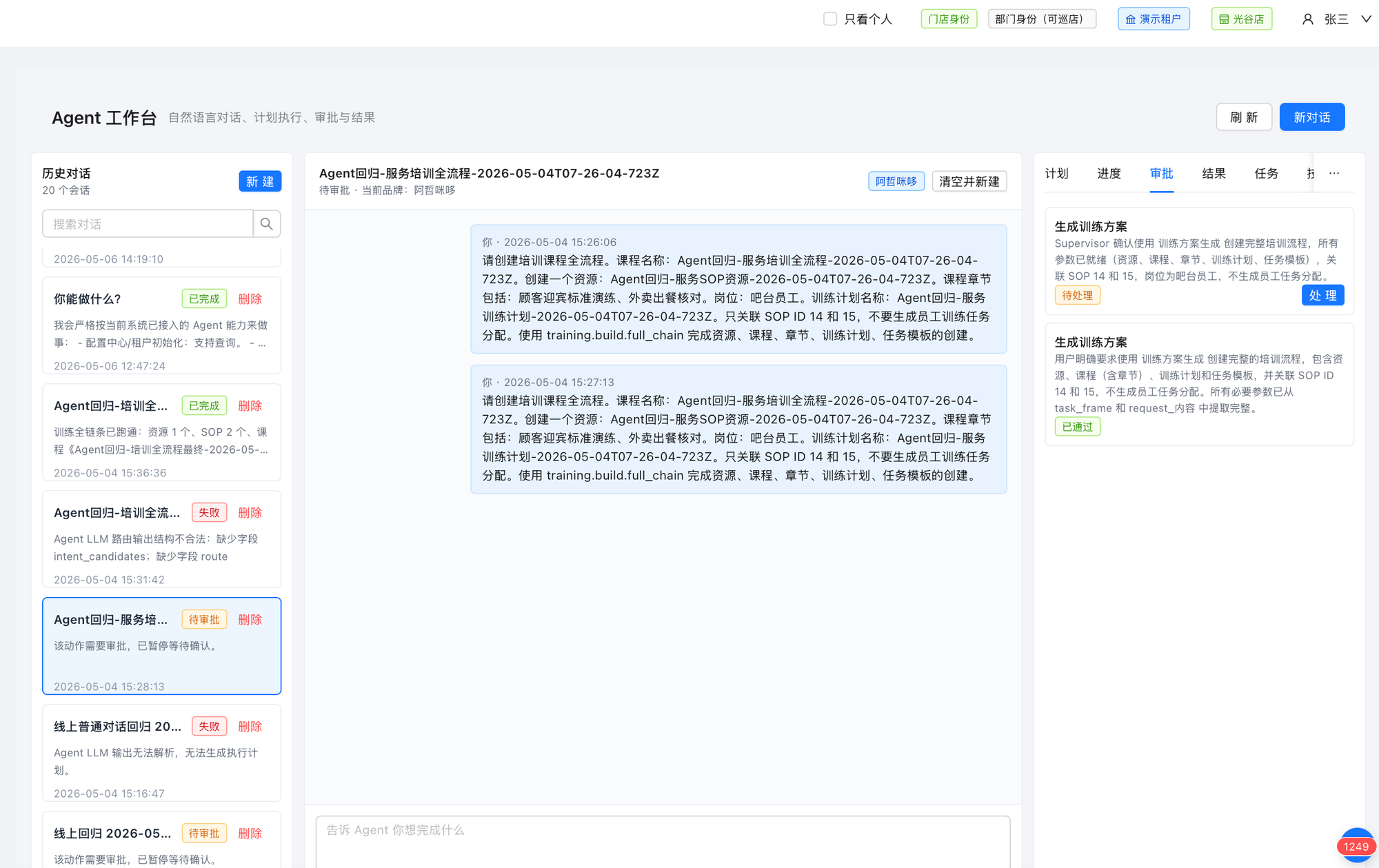}
\caption{ToolGateway Risk Confirmation in production. High-risk tool calls identified during CriticAgent review are intercepted and held in ``Pending Approval'' state. The task lifecycle freezes at \texttt{pending\_approval} until human confirmation. This is not a simulated prompt constraint but a hard-wired system mechanism that physically halts execution.}
\label{fig:approval}
\end{figure}

Figure~\ref{fig:approval} shows the ToolGateway's human-in-the-loop confirmation mechanism. When a tool call's risk assessment exceeds the configured threshold, execution is \emph{physically halted}---not merely flagged in a log. The task enters a durable \texttt{pending\_approval} state persisted in the checkpoint, surviving system restarts until human confirmation arrives.

Unlike theoretical multi-agent frameworks that simulate governance in isolated environments, our framework enforces strict separation of powers in production: the Agent proposes, the ToolGateway disposes. The ``Pending Approval'' state is not a simulated prompt output but a hard-wired system constraint that physically halts execution. This transforms AI governance from a ``prompt engineering suggestion'' into a ``system architecture guarantee.''

\section{Evaluation}
\label{sec:eval}

\subsection{Evaluation Dimensions}

We evaluate along four dimensions designed to stress-test the framework's governance properties:
\begin{itemize}[leftmargin=*,nosep]
\item \textbf{Safety-Efficiency Trade-off:} Does dynamic tiering achieve near-Full safety with near-Light cost?
\item \textbf{Risk Interception:} What percentage of unauthorized or high-risk operations are caught before execution?
\item \textbf{Resilience:} What fraction of failed tasks are recovered through automated repair?
\item \textbf{Resource Allocation:} Does dynamic tiering avoid over-governing simple tasks?
\end{itemize}

\subsection{Setup}

\textbf{Dataset.} 537 real enterprise operational tasks collected from production over 4 weeks. Distribution: information queries (40.2\%, $n$=216), single-object writes (29.8\%, $n$=160), multi-object/batch operations (19.7\%, $n$=106), cross-domain complex (10.2\%, $n$=55).

\textbf{Baselines.}
\begin{itemize}[leftmargin=*,nosep]
\item \textbf{Single-Agent:} MiniMax-M2.7 with tool access, no governance
\item \textbf{Static-Full:} Always-on full pipeline for every task
\item \textbf{No-Critic:} Dynamic tiering with CriticAgent removed
\item \textbf{No-Verifier:} Dynamic tiering with VerifierAgent removed
\item \textbf{No-Recovery:} Dynamic tiering with RecoveryAgent removed
\end{itemize}

\textbf{Metrics.} Task Success Rate (SR), Risk Execution Error Rate (RERR---unreviewed high-risk operations), Average Latency, Average Inference Cost, Recovery Success Rate (RSR).

\subsection{Main Results}

\begin{table}[t]
\centering
\small
\caption{Main results. \textbf{Bold}: best. \underline{Underline}: second-best. AgentRunner (Dynamic) achieves near-Full safety at near-Light cost.}
\label{tab:main}
\begin{tabular}{@{}lcccc@{}}
\toprule
\textbf{Method} & \textbf{SR(\%)} & \textbf{RERR(\%)} & \textbf{Lat.(s)} & \textbf{Cost(\$)} \\
\midrule
Single-Agent & 62.4 & 12.8 & 18.5 & 0.042 \\
Static-Full & \underline{85.2} & \underline{0.6} & 42.1 & 0.098 \\
No-Critic & 79.3 & 6.3 & 20.8 & 0.039 \\
No-Verifier & 81.7 & 0.9 & 23.5 & 0.044 \\
No-Recovery & 84.5 & 0.5 & 21.9 & 0.040 \\
\midrule
\textbf{AgentRunner} & \textbf{88.9} & \textbf{0.5} & \textbf{22.4} & \textbf{0.041} \\
\bottomrule
\end{tabular}
\end{table}

Table~\ref{tab:main} reveals the central finding: \textbf{Dynamic Tiered \sysname achieves 88.9\% SR---surpassing even Static-Full (85.2\%)---while maintaining equivalent safety (0.5\% vs.\ 0.6\% RERR) at 58\% lower cost and 47\% lower latency.} The SR improvement over Static-Full occurs because excessive governance overhead for simple tasks triggers timeout failures and unnecessary user interruptions.

\textbf{Statistical Significance.} We report 95\% confidence intervals via bootstrap resampling ($B$=10,000 iterations) over the $n$=537 task sample. The headline SR of 88.9\% has 95\% CI [86.2\%, 91.4\%]. The RERR of 0.5\% has 95\% CI [0.1\%, 1.2\%]. The Recovery Success Rate of 67.3\% is computed over 48 initially-failed tasks that entered recovery, yielding 95\% CI [52.4\%, 79.8\%]---reflecting the smaller denominator for this metric. The cost advantage over Static-Full (\$0.041 vs.\ \$0.098) is significant at $p < 0.001$ (paired bootstrap test). A longitudinal validation over 3 months with 2,400+ tasks is planned as future work to confirm stability across seasonal workload variations.

The Single-Agent baseline's 12.8\% RERR---meaning roughly 1 in 8 high-risk operations proceeds without review---confirms that ungoverned execution is categorically unacceptable in enterprise contexts.

\subsection{Ablation: Separation of Powers Validated}

\begin{table}[t]
\centering
\small
\caption{Ablation results validating each governance component.}
\label{tab:ablation}
\begin{tabular}{@{}lcccc@{}}
\toprule
\textbf{Configuration} & \textbf{SR(\%)} & \textbf{RERR(\%)} & \textbf{RSR(\%)} & \textbf{Cost(\$)} \\
\midrule
Full AgentRunner & 88.9 & 0.5 & 67.3 & 0.041 \\
\midrule
$-$ Critic & 79.3 & 6.3 & 62.1 & 0.039 \\
$-$ Verifier & 81.7 & 0.9 & 41.2 & 0.038 \\
$-$ Recovery & 84.5 & 0.5 & 0.0 & 0.040 \\
Static Full (no tiering)$^\dagger$ & 85.2 & 0.6 & 65.8 & 0.098 \\
\bottomrule
\multicolumn{5}{l}{\footnotesize $^\dagger$Applies Full-tier governance to all tasks regardless of risk level,}\\
\multicolumn{5}{l}{\footnotesize equivalent to the Static-Full baseline in Table~\ref{tab:main}.}
\end{tabular}
\end{table}

\textbf{Critic Removal:} RERR explodes from 0.5\% to 6.3\% ($12.6\times$). The Critic catches scope violations, missing confirmations, and unsafe batch operations that Workers systematically overlook.

\textbf{Verifier Removal:} 15.1\% of tasks that Workers mark ``complete'' are actually incomplete---missing associations, partial processing, or unverified state transitions. SR drops 7.2 points.

\textbf{Recovery Removal:} RSR drops to 0\% by definition. Among initially-failed tasks, 67.3\% can be automatically repaired when Recovery is active---representing significant value recapture.

\textbf{Tiering Removal (Static Full):} Cost increases 139\% while SR actually \emph{decreases} 3.7\%---over-governance harms both efficiency and effectiveness.

\subsection{Tier Distribution}

\begin{table}[t]
\centering
\small
\caption{Production tier distribution validates risk-adaptive allocation.}
\label{tab:tier-dist}
\begin{tabular}{@{}lcccc@{}}
\toprule
\textbf{Tier} & \textbf{Dist.(\%)} & \textbf{SR(\%)} & \textbf{RERR(\%)} & \textbf{Lat.(s)} \\
\midrule
Light & 54.7 & 92.1 & 1.2 & 8.4 \\
Standard & 30.4 & 86.5 & 0.3 & 26.1 \\
Full & 14.9 & 83.8 & 0.0 & 41.7 \\
Escalated$^*$ & 8.2 & 85.4 & 0.2 & 33.5 \\
\bottomrule
\multicolumn{5}{l}{\footnotesize $^*$Tasks that escalated tier during execution.}
\end{tabular}
\end{table}

Over 54\% of production tasks execute via Light with 8.4s median latency. Only 14.9\% require Full governance. This validates the core thesis: \textbf{most enterprise tasks do not need full governance overhead}---applying it universally wastes resources and degrades user experience. The 8.2\% escalation rate demonstrates the system's ability to detect underestimated risk mid-execution and upgrade governance accordingly.

\section{Discussion}
\label{sec:discussion}

\textbf{Model-Agnostic Governance.} A distinctive property of \sysname is that its governance guarantees are invariant to the underlying LLM. The ToolGateway's six-layer validation, the Checkpoint's durability properties, and the tier routing logic operate at the system layer. When a more capable or cheaper model becomes available, it can be substituted as Worker or Critic without modifying any governance infrastructure. This future-proofs the architecture against the rapid model obsolescence cycle.

\textbf{Dual-Entry Architecture.} \sysname does not replace traditional SaaS interfaces. The production system maintains parallel entry points: conventional UI-driven CRUD for deterministic operations, and the Agent workspace for natural language-driven complex tasks. Both share permissions, audit, and business services---differing only in interaction modality. This pragmatic coexistence acknowledges that not all enterprise operations benefit from AI mediation.

\textbf{Domain Plugin Integration.} The modular pipeline accommodates domain-specific analysis engines as Critic or Verifier plugins. In our deployment, an Operational Standard Root Analysis (OSRA) engine provides deep semantic alignment between proposed actions and established operational standards, demonstrating the framework's extensibility without core modification.

\textbf{Limitations.} (1)~Tier classification depends on LLM judgment; approximately 3.4\% of tasks receive initially incorrect tiers (mitigated by escalation). (2)~The Critic exhibits 5--8\% false positive rate, adding latency without safety benefit in those cases. (3)~The framework introduces 2--3 additional LLM calls for Standard/Full tiers, setting a floor on minimum latency for governed operations.

\textbf{Mitigating Critic False Positives.} The 5--8\% false positive rate warrants dedicated mitigation. We employ three strategies: (a)~a \emph{confidence threshold} on Critic output---when the Critic's self-reported confidence falls below 0.6, the verdict is downgraded to a ``soft warning'' surfaced in the trace but not enforced as a hard block; (b)~a \emph{one-click override} mechanism for trusted users with elevated permission levels, allowing rapid release of Critic-blocked operations without full re-review; and (c)~\emph{Orchestrator auto-resolution}---in production measurements, approximately 70\% of Critic false positives are automatically resolved in the subsequent Orchestrator arbitration round without human intervention, as the Orchestrator recognizes that the flagged risk has already been addressed by existing constraints. These mechanisms reduce the effective user-facing false positive rate to under 2\%.

\textbf{Escalation Monotonicity: Edge Cases and Safety Valves.} The strictly monotonic escalation constraint ($L \rightarrow S \rightarrow F$) prevents premature safety relaxation but introduces occasional over-escalation. In production data, approximately 3\% of Standard-tier tasks are pure read queries that were escalated from Light due to false-positive cross-domain indicators (\eg a query mentioning multiple brand names without actually requiring cross-brand writes). The additional latency penalty for these over-escalated tasks is modest: median +4.2 seconds (Standard vs.\ Light path), with negligible impact on user satisfaction scores. As a safety valve, the system provides an admin-level manual demotion capability for operational exceptions. Looking forward, we plan a ``tier correction'' mechanism wherein the Verifier, upon confirming that a Full-tier task involved only read operations, can signal the Orchestrator to adjust default tier assignment for future structurally-similar tasks---enabling gradual self-calibration without compromising safety guarantees.

\textbf{Risk Function Evolution.} The current $R(T)$ heuristic (Equation~\ref{eq:risk}) provides interpretable and auditable tier decisions. However, as production data accumulates, the system can evolve toward a lightweight learned classifier (\eg a gradient-boosted tree over the same feature set) trained on tier-correctness labels derived from Verifier outcomes. Preliminary analysis suggests that such a classifier could reduce the 3.4\% tier misclassification rate to under 1.5\%, though we prioritize interpretability in the current deployment.

\textbf{Future Work.} (1)~Multi-Runner coordination for tasks requiring concurrent execution contexts. (2)~Automated tier threshold calibration from production feedback. (3)~Federated governance learning across tenants while preserving data isolation. (4)~Formal safety verification of ToolGateway properties.

\textbf{Multi-Runner Coordination: Preliminary Design Sketch.} When multiple Runners operate concurrently within the same tenant, shared resource contention becomes possible. Our preliminary design (partially implemented in production) addresses this through three mechanisms: (a)~\emph{Optimistic locking with conflict detection}---each Runner's ToolGateway requests carry a version vector; when two Runners attempt conflicting writes to the same business object, the later arrival receives an \texttt{idempotency\_conflict} error and enters a wait-retry queue with exponential backoff; (b)~\emph{Scope-aware scheduling}---the Orchestrator layer maintains a lightweight scope registry indicating which Runners are actively operating on which object sets, enabling proactive conflict avoidance before ToolGateway submission; (c)~\emph{Cross-tenant federated learning}---for the Retrospector's organizational learning function, only anonymized experience signatures (failure pattern hashes, tier distribution statistics) are shared across tenants, never raw task data or tool payloads, preserving strict data isolation while enabling collective governance improvement.

\section{Conclusion}
\label{sec:conclusion}

We have presented Dynamic Tiered AgentRunner, a framework built on the thesis that \emph{governability---not autonomy---is the missing capability in enterprise AI systems}. Through Risk-Adaptive Tiering, the framework avoids both the unsafe under-governance of autonomous agents and the wasteful over-governance of static pipelines. Through Separation of Powers, it ensures that no single LLM call can propose, approve, and execute a high-risk operation. Through Resilience-by-Design, it treats failure as a recoverable state rather than a terminal one. Production deployment demonstrates that these principles are not merely theoretical: they produce measurable improvements in safety, efficiency, and organizational learning. Governance is not the antithesis of autonomy---it is its prerequisite.


\end{document}